\documentclass[letterpaper, 10 pt, conference]{ieeeconf} 
\IEEEoverridecommandlockouts
\usepackage{cite}
\usepackage{amsmath,amssymb,amsfonts}
\usepackage{algorithmic}
\usepackage{graphicx}
\usepackage{textcomp}
\usepackage{xcolor}
\usepackage{subfigure}
\usepackage{tablefootnote}

\newcommand{\R}{\mathbb{R}}

\begin{document}

\title{A Comparative Study of Using Spatial-Temporal Graph Convolutional Networks for Predicting Availability in Bike Sharing Schemes\\}

\author{Zhengyong Chen, Hongde Wu, Noel E. O'Connor and Mingming Liu $^{\star}$% <-this % stops a space
	\thanks{Z. Chen, H. Wu, N. O'Connor, and M. Liu are with the School of Electronic Engineering, Dublin City University, Dublin, Ireland. Z. Chen and H. Wu are joint first authors.}%
	\thanks{$^{\star}$Corresponding author. Email: {\tt mingming.liu@dcu.ie}}%
}

\maketitle

\begin{abstract}
Accurately forecasting transportation demand is  crucial for efficient urban traffic guidance, control and management. One solution to enhance the level of prediction accuracy is to leverage graph convolutional networks  (GCN), a neural network based modelling approach with the ability to process data contained in graph based structures. As a powerful extension of GCN, a spatial-temporal graph convolutional network (ST-GCN) aims to capture the relationship of data contained in the graphical nodes across both spatial and temporal dimensions, which presents a novel deep learning paradigm for the analysis of complex time-series data that also involves spatial information as present in transportation use cases. In this paper, we present an Attention-based ST-GCN (AST-GCN) for predicting the number of available bikes in  bike-sharing systems in cities, where the attention-based mechanism is introduced to further improve the performance of an ST-GCN. Furthermore, we also discuss the impacts of different modelling methods of adjacency matrices on the proposed architecture. Our experimental results are presented using two real-world datasets, Dublinbikes and NYC-Citi Bike, to illustrate the efficacy of our proposed model which outperforms the majority of existing approaches.
\end{abstract}

%\begin{IEEEkeywords}
%ST-GCN, Attention-based Model, Adjacency Matrix, Bike Sharing, Availability Prediction 
%\end{IEEEkeywords}

\section{Introduction}

Nowadays, there is increasing interest and demand for adopting bike-sharing systems globally. In fact, an efficient bike-sharing system can not only reduce cost and commute time for urban commuters, but can also effectively mitigate the level of air pollution emissions generated in cities \cite{otero2018health}. An important consideration to make bike-sharing system efficient is to  balance supply and demand in the bike-sharing network \cite{raviv2013optimal}. To do this, traditional management methods such as manual monitoring systems, have been deployed to enable relocation of bikes across different stations using other means of transportation, e.g. trucks. However, this approach can easily lead to supply-demand imbalance due to estimation errors of system operators and unexpected traffic delays during the bike transition. Thus, due to the uncertainty of departure and arrival of bikes at any bike station, it is important to take a more proactive approach by accurately predicting the number of  bikes that will be available for users to access at any given time and location.

Recently, convolutional neural networks (CNN) have been applied to extract the relationship between adjacent traffic networks whilst the recurrent neural networks (RNN) were used to arrest the temporal information. For short-term traffic prediction, fully connected long short-term memory (LSTM) \cite{shi2015convolutional} and CLTFP \cite{wu2016short}, two architectures mixed the long short-term memory networks with convolutional operation, were proposed in order to catch both temporal and spatial cues. However, LSTM or other networks with recurrent architecture are computationally intensive and hard to converge the network parameters with global optimization, since the recursive training accumulates the error for the prediction. On the other hand, CNN-based methods also have their limitation since the convolution process the data in 2-D form restrictively, which may not be the natural structure of traffic data.

These above issues of CNN and RNN-based methods were investigated and addressed by the spatial-temporal graph convolutional networks (ST-GCN) \cite{DBLP:conf/ijcai/YuYZ18}, a variant of a graph neural network (GNN) for utilizing spatial information. Spatial-temporal convolutional blocks were introduced and applied repeatedly in this architecture, combining several graph convolutional layers \cite{DBLP:conf/nips/DefferrardBV16} with sequential convolution in order to  represent the spatial-temporal relations. Subsequent to this approach, STG2Seq \cite{bai2019stg2seq}, a sequence-to-sequence variant of STGCN, is proposed with more reference on historical data and an attention module, for multi-step passenger demand forecasting. However, there are still some important issues to be solved in the ST-GCN architecture. For instance, how effective a specific adjacency matrix scheme can contribute to traffic demand prediction. Also, to what extent an attention-based mechanism can be applied to further improve the accuracy for a given demand prediction model. 

To answer these questions, our key objective in this paper is to investigate how ST-GCN, supplemented with an attention-based mechanism, can further enhance the performance of bike availability prediction across different bike stations in cities. From an application/service perspective, we believe the proposed method can help cyclists make their personalized travel plan more appropriately by finding the best bike station nearby with high confidence in availability. Thus, the contribution of our work can be summarized as follows: 

\begin{itemize}
    \item[1.] We combine an attention mechanism with the ST-GCN, namely AST-GCN, to improve the ability of extracting spatial-temporal features for the prediction task. In comparison with the existing methods, our model shows a promising performance. 
    \item[2.] We review related works in the recent literature and summarize four categories for modelling adjacency matrices, namely spatial based, temporal based, spatial-temporal based and adaptive based adjacency matrix.
    \item[3.] Given our findings in 1 and 2, we evaluate our proposed AST-GCN model with the adjacency matrices of interest using a real-world dataset, Dublinbike, for bike sharing availability prediction. Our results show that: (a) adaptive spatial-temporal adjacency matrix can achieve the best performance; (b) spatial-temporal based adjacency matrix can achieve better results than that only using spatial-based or temporal-based adjacency matrix; (c) spatial-based adjacency matrix achieves similar performance as the temporal-based one.
\end{itemize}

The rest of the paper is organized as follows. We introduce some previous researches related to traffic demand prediction in Section \ref{RW} and formulate our problem in Section \ref{Method}. Experimental setups are demonstrated in Section \ref{experiment} and the results are discussed in Section \ref{result}. Finally, we summarize our work in Section \ref{concl}.

\section{Related Work} \label{RW}

\subsection{Existing Methods}
In general, forecasting traffic demand is difficult, when a traffic demand depends not only on the historical demand pattern of the target area (e.g., suburb) but also on the pattern of other areas (e.g., urban). To meet this challenge, many studies using deep learning such as CNN, RNN, and GNN have been proposed.

As the traditional convolutional operation in CNN process the data with a 2D approach, the layout of a city is geographically divided into square blocks in order to extract spatial relationships from all regions \cite{zhang2017deep}, nearest regions \cite{yao2018deep} or in other 2D forms \cite{chu2020passenger}. RNN based methods and their variants \cite{yao2019revisiting} are applied to catch temporal correlation, for instance, structuring the historical traffic demand sequence for each region \cite{shi2015convolutional} and presented as a 1D feature-level fused architecture \cite{wu2016short}. GNN based methods, with natural advantages in utilizing spatial information, model the traffic network by a general graph instead of treating the traffic data arbitrarily (e.g., grids and segments) in CNN and RNN methods. GCN, as a variant of GNN, which is able to combine spatial and temporal information, is widely used in the scenario of traffic demand prediction as seen in many recent works \cite{DBLP:conf/nips/DefferrardBV16} \cite{bai2019stg2seq} \cite{DBLP:conf/ijcai/YuYZ18}.

Attention is a popular technique in deep learning that mimics physiological cognitive attention. The effect enhances the importance of small parts of the input data and de-emphasising the rest. This technique has been used to enhance the prediction performance for many sequence-based tasks of GNNs, i.e. Graph attention networks \cite{velivckovic2017graph}. In traffic demand prediction, the importance of each previous step to target demand is different, and this influence changes with time. For instance, a temporal attention mechanism \cite{bai2019stg2seq} is able to add an importance score for each historical time step to measure the influence and this strategy can effectively improve the accuracy on prediction accuracy.

\subsection{Adjacency Matrices} \label{AM}
An adjacency matrix is used to indicate whether a pair of vertices is connected by edge or not in graph data. For a traffic network, it is important to understand how an adjacency matrix can be used to best capture the interconnectivity between different nodes in the graph. To the best of our knowledge, four types of adjacency matrices have been investigated in  previous research works, namely spatial (S), temporal (T), spatial-temporal (ST) and adaptive (A). A spatial adjacency matrix is usually distance-based. Euclidean distances between different stations (i.e., nodes in graph) \cite{DBLP:conf/ijcai/YuYZ18}  \cite{chen2020multitask} or the natural geographical distance \cite{kim2019graph} are usually used as weights for its entries. A temporal adjacency matrix can be defined based on the similarity score \cite{bai2019stg2seq} (i.e., Pearson correlation coefficient) between the temporal information (i.e., historical traffic demand sequence) of each pair of nodes/stations. To combine the benefits of both spatial and temporal features, an spatial-temporal embedding (ST embedding) can be generated for each node in a graph \cite{ye2020coupled}. However, in such a scenario, it can be hard to describe the adjacency matrix intuitively with the high dimension embedding features and thus the adjacency matrix needs to be adaptively defined along with the training process of GCN \cite{wu2019graph} \cite{chiang2019cluster}.

\section{Methodology}\label{Method}

\subsection{Notations and Problem Statement}

We consider a scenario where $N$ bikes stations are included as part of a bike-sharing system. Let $\underline{\textbf{N}}:= \left\lbrace 1,2,\ldots, N \right\rbrace$ be the set for indexing the bike stations in the system. For a given bike station $i \in \underline{\textbf{N}}$, let $A^i_t \in \R$ be the number of available bikes at the station $i$ at time $t$. We denote $\textbf{A}_t \in \R^{N}$ the vector consisting of the number of available bikes across all stations $N$ at time $t$. In addition, each bike station $i$ is associated with a set of features for model training, e.g. weather condition, weekday, etc, and let  $F^i_t \in \R^{d}$ represent the values of its features at time $t$, where $d$ is the number of features used. Similarly, we let $\textbf{F}_t \in \R^{N \times d}$ be the feature set values of all bike stations at time $t$. Given the notation above, our learning objective is to find a function $\textbf{H}(.)$ which is able to address the following problem:
\begin{equation*}
\textbf{A}_{t+1:t+n} = \textbf{H}(\textbf{A}_{t-m+1:t}; \textbf{F}_{t-m+1:t})
\end{equation*}

\noindent where $m, n$ denotes the input and output length for the model respectively. Also, the notation ${t+1:t+n}$ presents the output as a sequence of vectors from steps $t+1$ to $t+n$.

\subsection{Attention-based ST-GCN}

In this section, we introduce the attention-based ST-GCN architecture that  used for solving our bike sharing availability prediction problem. We note that the ST-GCN architecture has been presented in \cite{DBLP:conf/ijcai/YuYZ18}, and the architecture consists of two identical ST-Conv-Blocks and a fully connected output layer. Specifically, an ST-Conv-Block consists of two temporal gated convolutional (TGC) layers and one spatial graph convolutional (SGC) layer, which are the essential modules of ST-GCN. In general, TGC is in charge of extracting temporal features and SGC is able to extract spatial features from the data. However, since there is no attention on the temporal channel of ST-GCN, this significantly degrades the performance for sequence to sequence based learning tasks. As such,  the model's learning capability may be significantly reduced due to ``lost of focus''. To deal with this issue, we introduce a temporal-attention module (TAM) in each ST-Conv-Block, as shown in Fig. \ref{fig:stgcn} where the temporal-attention module is depicted in green.  \\

\noindent \textbf{Remark:} An attention mechanism was introduced in \cite{shiraki2020spatial} and \cite{zhang2020sta} to extract both spatial and temporal information from ST-GCN networks. The architectures proposed in both works applied attention operation to extract spatial and temporal information separately. In particular, the model in \cite{shiraki2020spatial} consisted of 15 ST-Conv blocks in total with two attentions matrices calculated from them, while the model in \cite{zhang2020sta} was stacked by 10 ST-Conv blocks with two attention matrices computed from each ST-Conv block. With increased model complexity and computation cost, stacking multiple ST-Conv blocks with attention matrices calculated separately may be of less interest since the spatial and temporal information may not be combined towards an effective spatial-temporal embedding in such a case. Instead, our model only consists of 2 ST-Conv blocks and the proposed AST-GCN architecture lightly merges spatial-temporal information with attention by calculating the attention matrix only once in each ST-Conv Block, which reduces the computation costs during the model training process. Specifically, the first TGC module generates original temporal information and the last TGC module generates spatial-temporal information (as it takes account of the output of the preceding SGC layer as its input). Passing through two average 3D pooling layers, both information are combined before a Relu activation function is applied. A sigmoid function is connected here to generate probabilistic weights (attention matrix) with values between 0 and 1. With this matrix in place, the attention-based temporal information is generated by using a dot product with the output of the first TGC layer and then concatenated as input to the subsequent ST-Conv Block. Both spatial and temporal information in the data flow are fully captured before passing to the dense layer for sequential output prediction. 
\vspace{-0.3cm}

\begin{center}
	\begin{figure*}[htbp]
		\centering
		\includegraphics[width=5.5in, height=2.6in]{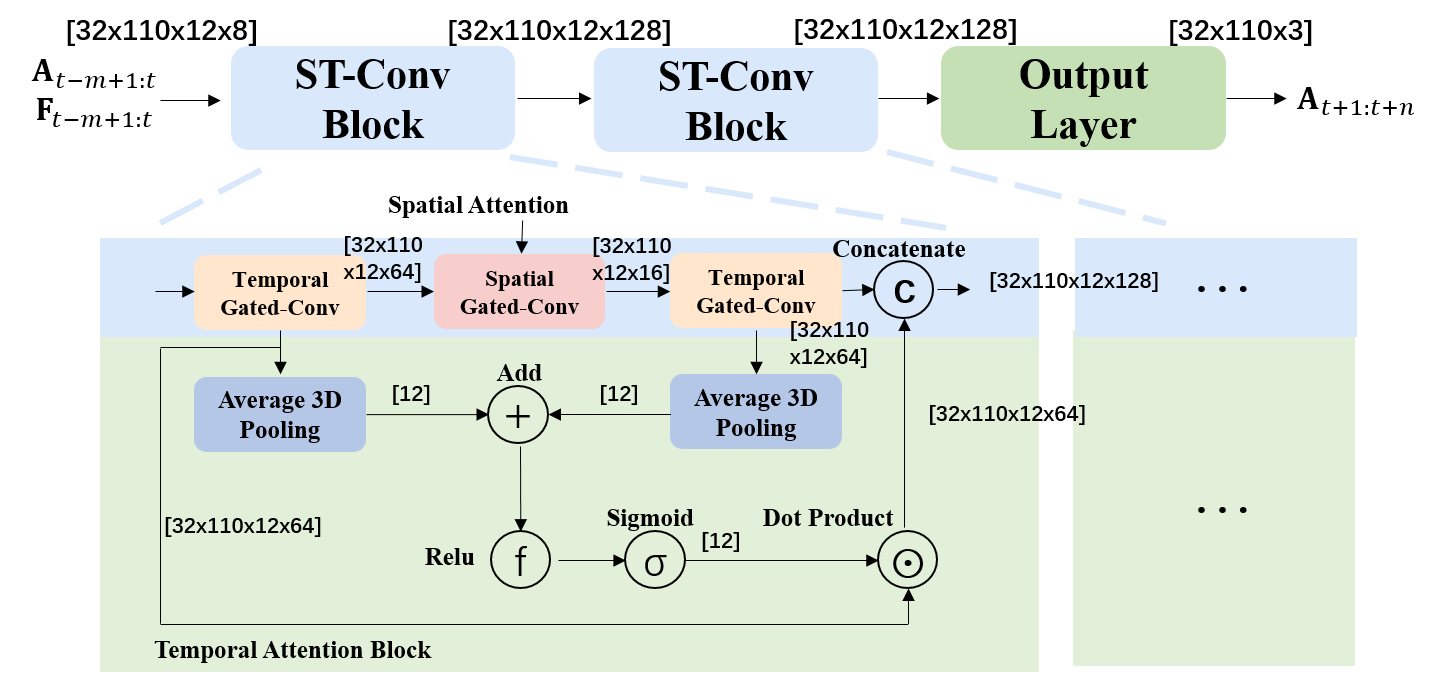}
		\caption{The proposed ST-GCN with a temporal-attention module (TAM).}\label{fig:stgcn}
	\end{figure*} \vspace{-0.1cm}
\end{center}

\section{Algorithms and Experiments} \label{experiment}

In this section, we discuss the different configurations investigated for comparative studies. 

\subsection{Experimental Datasets}
\begin{figure}[htp]
	\centering
	\includegraphics[width=8cm, height=1.25in]{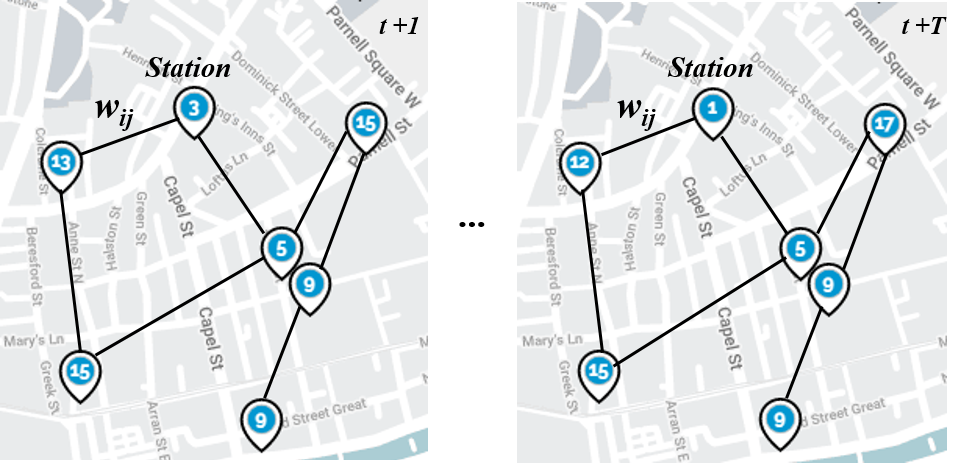}
	\caption{A subset of bike stations of the Dublin bike sharing system operates in real-time. The edges are added for the illustration of the inherent graph signals.}
	\label{fig:bike}
\end{figure}

%Fig. \ref{fig:1}
\begin{itemize}
    \item Dublinbike:
    DublinBikes is a bike-sharing scheme in operation in Dublin City, Ireland. The system is illustrated in Fig. \ref{fig:bike}, where each node is a bike station and each blue number in the circle indicates the number of available bikes in real-time. Real-time data is accessible using an API and we also have access to historic data, recorded every five minutes, which includes timestamps, station states, number of available bikes and station locations, etc. We choose the data \footnote[1]{https://data.smartdublin.ie/dataset/analyze/33ec9fe2-4957-4e9a-ab55-c5e917c7a9ab} from 01/07/2020 to 01/10/2020 for our studies.
    \item NYC-Bikes\cite{2016DNN}: This dataset includes the NYC Citi daily bike orders of people  using the bike sharing scheme. We choose the transaction records from April 1st, 2016 to June 30th, 2016 (91 days). This contains the following information: bike pickup station, bike drop-off station, bike pick-up time, bike drop-off time and trip duration.
    \item Visualcrossing Weather Data \footnote[2]{https://www.visualcrossing.com/weather-data}: This dataset provides weather conditions at different locations at different historical time points, including temperature, humidity and wind speed, etc. This weather dataset has been integrated for experiments that use  the Dublinbikes dataset.
\end{itemize}

\subsection{Experimental Setup}
\begin{itemize}
	\item Dublinbike: 
	For this scenario, we use the number of available bikes at each bike station in the first 3 hours to predict the number of available bikes at each bicycle station 45 minutes later, where each data point is the averaged number of available bikes in 15 minutes. This implies that we take the past 12 consecutive observation points to predict the following 3 points of our interest. The dataset consists of 110 bike stations in total. The data is then separated into a training set (60$\%$), a validation set (20$\%$) and a testing set (20$\%$) in a sequential manner.
	\item NYC-Bikes:
	NYC Citi Bike is dock-based and every depot of bikes is considered as a station. Following the same experiment setup as in CCRCN \cite{ye2020coupled}, we filter out the stations with fewer orders and keep the 250 stations with the most orders. The time step is set to half an hour. Among the last four weeks considered, the first two are used for validation, and the last two are for testing.
\end{itemize}

To evaluate the performance across different models, Mean Absolute Error (MAE) has been selected as the performance metric, indicating an intuitive margin between the predicted and the true amount of available bikes at each station.

\subsection{Baseline Algorithms}
\begin{itemize}
    \item Dublinbike:
To the best of our knowledge, there has been no GNN based methods implemented for the Dublinbike dataset. In particular, there has also been no ST-GCN based methods applied for solving the prediction for this dataset. For comparative studies, we conduct the experiments and use ST-GCN \cite{DBLP:conf/ijcai/YuYZ18} as our baseline.
    \item NYC-Bikes:
A lot of methods have been reported using this dataset to predict traffic demand. The state of the art work is presented in CCRCN \cite{ye2020coupled}. Based on this,  we compare the performance of different methods, including our proposed model, in a similar experimental setting. Specifically, the following methods are compared: (a) HA \footnote[3]{The average of historical values at previous time steps of a fixed length.}; (b) XGBoost\cite{xgboost2016}; (c) FC-LSTM\cite{lstm1997}; (d) DCRNN\cite{2017arXiv170701926L}; (e) ST-GCN\cite{DBLP:conf/ijcai/YuYZ18}; (f) STG2Seq\cite{bai2019stg2seq}; (g) GraphWaveNet\cite{wu2019graph} and (h) CCRNN\cite{ye2020coupled}.
\end{itemize}

\subsection{Network Setup}\label{net_set}

The historical data length used for both  the Dublinbikes dataset and the NYC-citi dataset is set to 12, the prediction length is set to 3 in Dublinbikes and 12 in NYC-citi respectively. The feature dimension used in NYC-citi is 2 representing the pick-up and drop-off demand. The feature dimension used for the  Dublinbikes dataset is 8, details of the feature selection will be discussed in the results section.  All models are optimized by Adam algorithm \cite{kingma2017adam}. Other setting of parameters  are presented in Table \ref{tab:Expri_settings}. The dimensions of the data flow during the training process of the proposed model are overlapped in Fig. \ref{fig:stgcn} for illustration purposes. It is worth noting that the input of the first temporal gated-Conv is strictly the same as the input of the corresponding ST-Conv block while the input of the second temporal gated-Conv is the output of previous spatial gated-Conv block. The concatenate operation concatenates the output of the first and the second temporal gated-Conv block. 
\begin{table}[htbp]
	\caption{Experiment Setting for two datasets } 
	\begin{center}
		\begin{tabular}{|l|c|c|c| }
			\hline
			\textbf{Setup} & \textbf{Dublinbikes} & \textbf{NYC-citi}\\
			\hline
			Station amount  & 110 &  250\\
			Historical data length  & 12 &  12\\
			Prediction length & 3 & 12\\
			Feature dimension & 8 & 2\\
			Batch size & 32 & 32\\
			Initial learning rates & 0.001 & 0.0001\\
			Optimizer & Adam algorithm & Adam algorithm\\
			Weight decay & 0.001 & N/A\\
			LR adjustment strategy & cosine annealing & adjust at equal intervals\\
			\hline
		\end{tabular}
	\end{center}
	\label{tab:Expri_settings}
\end{table}

\subsection{Adjacency Matrix Setup}
The adjacency matrix in the original ST-GCN architecture is not adjustable/trainable. As a result, this fixed adjacency matrix may not fully capture the spatial relationship between nodes in the graph. To improve it, we adapt the fixed adjacency matrix to a trainable adjacency matrix and then initialize the matrix using meaningful contextual information, e.g. distance between nodes, similarity between stations’ historical time-series data. Further, an adaptive adjacency matrix (AAM) is able to extract spatial attention information from the graph adaptively, and thus it makes the AST-GCN effective in capturing both spatial and temporal attention information. For our comparative studies, different setups of adjacency matrices are investigated as follows:

%On the other hand, this AAM can extract spatial attention information benefit from the fact that adjacency matrix decide the relationship between nodes. As we describe above, our proposed TAM also can learn the temporal attention information. So our proposed AST-GCN has the ability to learn spatial and temporal attention information. For the convenience of comparison, the adjacency matrices of different categories are modified on the basis of AST-GCN.

\begin{itemize}
    \item For the implementation of the adjacency matrix proposed in ST-GCN \cite{DBLP:conf/ijcai/YuYZ18}, the sigma is set to 0.2 and the epsilon is set to 0.368;
    \item For the implementation of the adjacency matrix proposed in STG2Seq\cite{bai2019stg2seq}, the sigma is set to 0.05;
    \item For the implementation of the adjacency matrix proposed in CCRCN\cite{ye2020coupled}, the dimension of station feature is set to 20 and the sigma is set to 1.
    \item Other adjacency matrices do not need parameters to be set. In other words, these adjacency matrices are calculated directly without parameters or are purely adaptive. 
\end{itemize}

\section{Results and Discussion} \label{result}
\subsection{Feature Selection for Dublinbikes Dataset}

In order to select the best features for our experiments, an ablation study has been carried out for a set of features which model temporal, spatial as well as weather characteristics. Specifically, we adopt ST-GCN as our basic setting for evaluation of different feature combinations. Our full feature sets are as follows: (1) number of available bikes (AB); (2) time of day (TD); (3) weekday (WD); (4) weather condition description (WCD); (5) temperature (T); (6) wind speed (WS); (7) cloud coverage (CC) and (8) Humidity (H). 

Results of the ablation study are reported in the Table \ref{tab:abl_result}, from which we easily conclude that the following feature combination gives the best performance: number of available bikes (AB), time of day (TD), weekday (WD) and weather conditions description (WCD). 

\begin{table}[htbp]
	\caption{Results of the ablation study of feature combinations}
	\begin{center}
		\begin{tabular}{|l|c|c|c| }
			\hline
			\textbf{Feature combination} & \textbf{MAE}\\
			\hline
			AB  & 3.24\\
			AB+TD & 3.19\\
			AB+TD+WD & 3.21\\
			\textbf{AB+TD+WD+WCD} & \textbf{3.16}\\
			AB+TD+WD+WCD+T & 3.36\\
			AB+TD+WD+WCD+WS & 3.40\\
			AB+TD+WD+WCD+CC & 3.30\\
			AB+TD+WD+WCD+H & 3.40\\
			All Together & 3.56\\
			\hline
		\end{tabular}
	\end{center}
	\label{tab:abl_result}
\end{table}

\subsection{Results Discussion}
\begin{itemize}
    \item NYC-Bikes:
    The results on the NYC dataset are compared between the proposed AST-GCN and existing algorithms reported in \cite{ye2020coupled} as shown in Table \ref{tab:nyc_stagcn}. It is shown that the AST-GCN algorithm outperforms the existing graph based architectures (i.e., ST-GCN and STG2Seq) with 24.67\% improvement in MAE, from 2.4976 to 1.8815. Also, although CCRNN beats all of its competitors, the AST-GCN shows minor difference in performance, and it still demonstrates comparable metrics compared to other sequence based models including Graph WaveNet, DCRNN.
    \item Dublinbikes:
    As shown in Table \ref{tab:stagcn}, after applying distance initialized AAM (DIAAM) on ST-GCN, the prediction results achieve better results with MAE equals 1.27. By replacing ST-GCN to AST-GCN, the MAE result has significantly improved from 1.27 to 1.04 for MAE. Among others, the embedding AAM (EAAM) makes the best performance which leads to the MAE equals 1. Results in Fig. \ref{fig:GTP} further highlight this key finding. Specifically, the biases between the ground truth and the first timestamp (i.e. NAB prediction for the first 15 minutes) as well as the third timestamp (i.e. NAB prediction for the 45 minutes) are both negligible showing that our proposed model can achieve impressive prediction performance for both short-term (15 mins) and long-term (45 mins) for the best case scenario.
\end{itemize}
\vspace{-0.75cm}

\begin{center}
	\begin{figure*}[ht]
		\centering
		\includegraphics[width=6.0in]{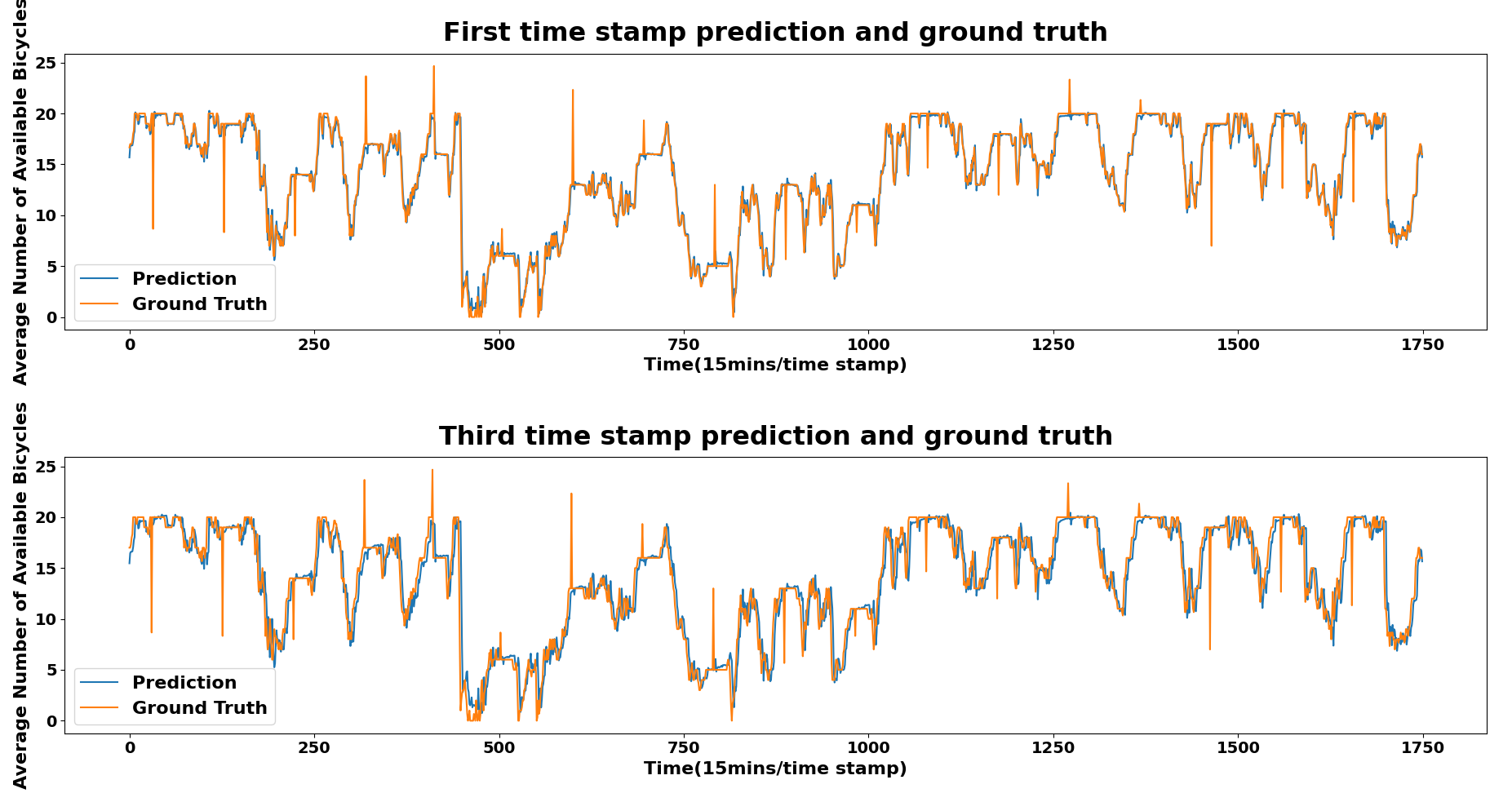}
		\caption{Comparison between ground truth and prediction.}\label{fig:GTP}
	\end{figure*}	\vspace{-0.1cm}
\end{center}

\begin{table}[htbp]
	\caption{Experiment result of AST-GCN on NYC-citi \cite{ye2020coupled}} 
	\begin{center}
		\begin{tabular}{|l|c|c|c| }
			\hline
			\textbf{Model} & \textbf{MAE}\\
			\hline
			HA & 3.4617\\
			ST-GCN & 2.7605\\
			STG2Seq & 2.4976\\
			XGBoost & 2.4690\\
			FC-LSTM & 2.3026\\
			Graph WaveNet & 1.9911\\
			DCRNN & 1.8954\\
			\textbf{AST-GCN + EAAM} & \textbf{1.8815}\\
			CCRNN & 1.7404\\
			\hline
		\end{tabular}
	\end{center}
	\label{tab:nyc_stagcn}
\end{table} 
\vspace{-0.5cm}

\begin{table}[htbp]
	\caption{Experiment result of AST-GCN on Dublinbikes} 
	\begin{center}
		\hspace{-0.5cm}
		\begin{tabular}{|l|c|c|c| }
			\hline
			\textbf{Model} & \textbf{Categories} \tablefootnote{The abbreviations in this column have been presented in Section \ref{AM}.} & \textbf{MAE (\%)} \\
			\hline
			ST-GCN + Euclidean distance& S & 1.36 (0\%) \\
			
			ST-GCN + DIAAM & S + A & 1.27 (-6.67\%) \\
			
			AST-GCN + DIAAM & S + A & 1.04 (-23.5\%) \\
			
			\textbf{AST-GCN + EAAM \cite{wu2019graph}} & ST + A & \textbf{1.00 (-26.5\%)}\\
			
			AST-GCN + Euclidean distance \cite{DBLP:conf/ijcai/YuYZ18}& S  & 1.06 (-22.0\%) \\
			
			AST-GCN + Geographical distance \cite{kim2019graph}& S & 1.09 (-19.8\%) \\
			
			AST-GCN + Temporal correlation \cite{bai2019stg2seq} & T & 1.07 (-21.3\%) \\
			
			AST-GCN + ST embedding \cite{ye2020coupled} & ST & 1.01 (-25.7\%) \\
			
			\hline
		\end{tabular}
	\end{center}
	\label{tab:stagcn}
\end{table}
\vspace{-0.5cm}

\subsection{Performance Evaluation w.r.t. Adjacency Matrices}

In this section, we discuss how different adjacency matrices can affect the learning performance for our proposed AST-GCN architecture. Our results are illustrated in Table \ref{tab:stagcn} where the percentage in parenthesis shows the difference of the achieved MAE in comparison to the basic setting: ST-GCN + Euclidean distance. Unsurprisingly, our results show that those fixed adjacency matrices, including both spatial based and temporal based, achieve the worst results among all other settings. In contrast, the adaptive-based settings can generally achieve better results compared to the fixed types, but with one exception for the spatial-temporal based setting, i.e. AST-GCN + ST embedding, which also shows a competitive result. For the adaptive-based settings, the embedding AAM, i.e. AST-GCN + EAAM, achieves the best result compared to the other AAM setting initialized by distance, i.e. AST-GCN + DIAAM. 

\subsection{Performance Evaluation w.r.t. Different Bike Stations} 

\begin{center}
	\begin{figure*}[htbp]
		\centering
		\includegraphics[width=5.4in]{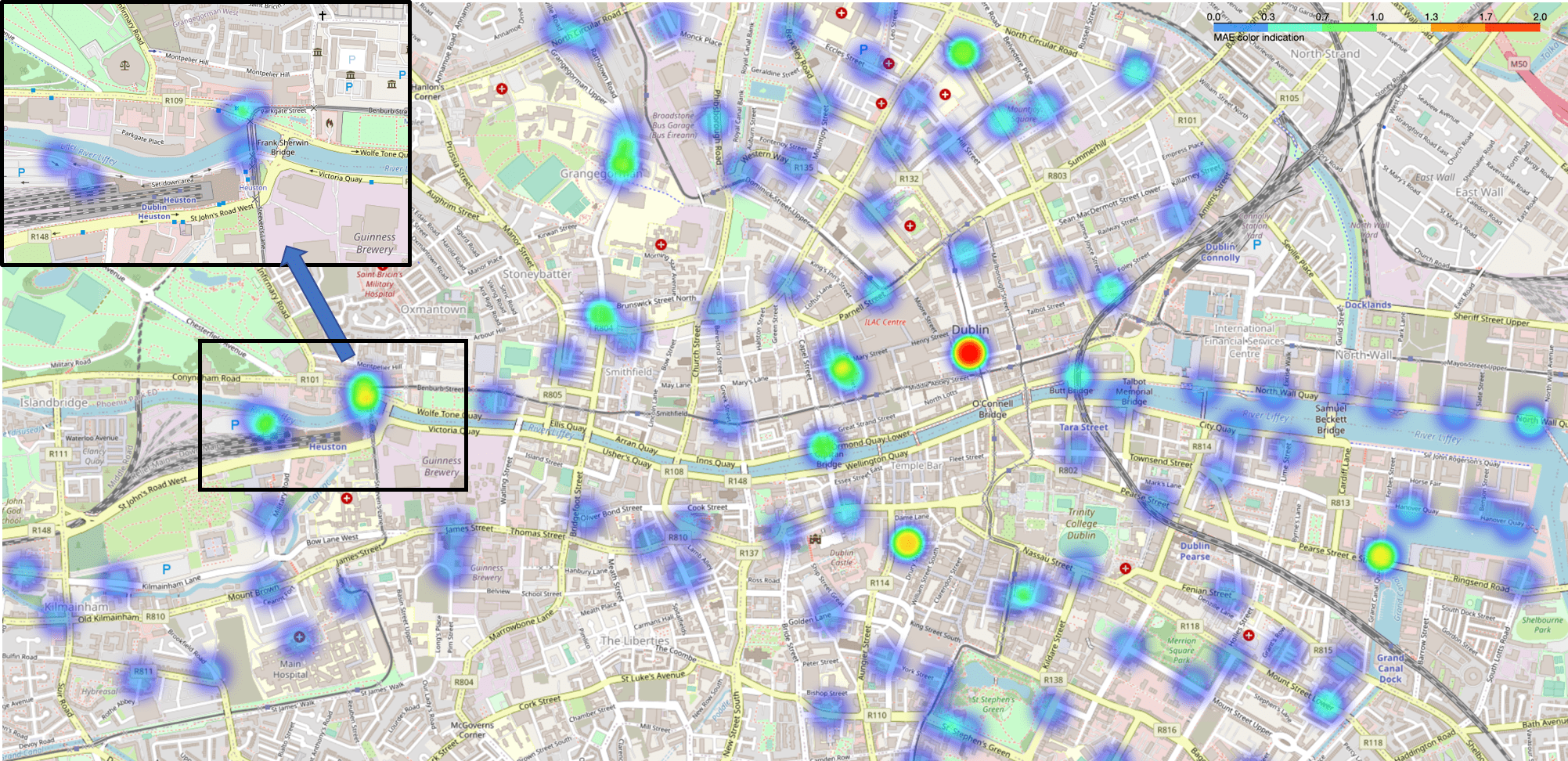}
		\caption{Heat-map of the station-wise MAE-based prediction error in Dublin city. A warm-toned color (red) indicates a higher MAE and a cool tone color (green-blue) indicates a lower MAE.}
		\label{fig:heatmap}
	\end{figure*}	
\end{center}

\begin{figure}[htbp]
	\centering
	\includegraphics[width=8.0cm]{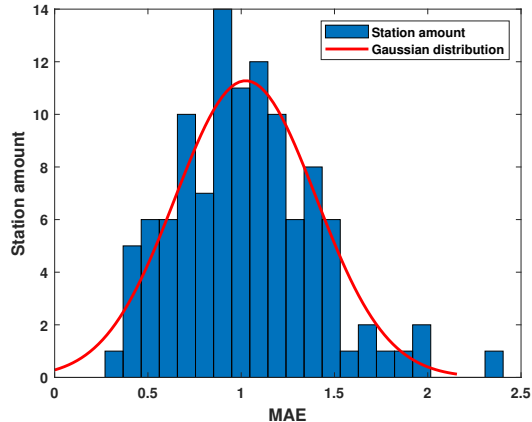}
	\caption{Histogram of station-wise MAE.}
	\label{fig:histogram}
\end{figure}

In this section, we present the prediction results for each bike station in the Dublinbike dataset using the best trained model (AST-GCN + EAAM). Our objective here is to illustrate the confidence with which a user can rely on our proposed prediction model to make a decision when he/she decides to get access to a bike from his/her nearby area. Our station-wise results are illustrated in Fig. \ref{fig:heatmap} and Fig. \ref{fig:histogram}. Specifically, Fig. \ref{fig:heatmap} shows the heat-map of station-wise MAE over the geographical map of Dublin city where the bike stations are facilitated. The red marks indicate a higher MAE and blue-green marks indicate a lower MAE in the corresponding area. Generally speaking, the results demonstrate that the prediction is more accurate (low-MAE values) outside of the city center showing that users can collect bikes with high confidence in the availability of bikes. The highest prediction error occurs in the heart of city centre, i.e. the bike station located at the ``Princes Street/O'Connell Street'', with the MAE equalling to 2.4. This may be caused by a frequent access and return of bikes by users in this central commuting area, leading to a relatively higher uncertainty in bike availability. The second highest prediction error appears in the western part of the city, i.e. the green-blue region indicated in the rectangular box in Fig. \ref{fig:heatmap}. However, this is mainly due to the aggregated effect where a few bike stations are very close to each other in the ``Benburb Street'' area. An in-depth view of the region, i.e., the upper left corner of the rectangular box in Fig. \ref{fig:heatmap}, further validates that the prediction error of each bike station therein is low. Finally, the statistical histogram of the station-wise MAE is illustrated in Fig. \ref{fig:histogram} showing that most bike stations have an MAE-based prediction error less than 1.5 bikes, which indicates that our proposed forecasting system is very robust and accurate for a number of bike stations in the Dublin city.

% \begin{center}
% \begin{figure}[ht]
% 	\centering
% 	\includegraphics[width=8cm]{heatmap_new.png}
% 	\caption{Heat-map of station-wise MAE in Dublin city. A warm-toned color (red) indicates a higher MAE while a cool tone color (blue) indicates a lower MAE. \HDcom{two column, change a new scroll view and add detail figure}}
% 	\label{fig:heatmap}
% \end{figure} \vspace{-0.1cm}
% \end{center}

\vspace{-0.75cm}

\section{Conclusion} \label{concl}

In this paper, we propose a spatial-temporal graph convolutional network architecture embedded with a temporal-attention module (AST-GCN) to predict the number of available bikes in bike-sharing systems  using realistic datasets. The temporal attention module is able to extract temporal attention information which aims to enhance the prediction accuracy compared to that of the original ST-GCN architecture reported in \cite{DBLP:conf/ijcai/YuYZ18}. Our experimental results show that the proposed AST-GCN can perform better than most of existing methods in the NYC-Citi dataset. As for the Dublinbikes dataset, our proposed model has demonstrated a very promising result of 1.00 MAE as the selected performance metric. In addition, we have thoroughly investigated how different modelling of the adjacency matrices can affect the overall model performance through a comprehensive comparative study on the DublinBikes dataset. Current results have shown that embedding AAM can achieve the best results compared to many other settings. 

To conclude, we believe that the work presented in this paper is an important step towards making  bike sharing systems more efficient thanks to the ST-GCN enabled techniques. Most importantly, we wish to note that it is our intention in this paper to deeply explore different variants of ST-GCN by using open access datasets to form a performance benchmark for the benefit of the research community. Further improvements to the network structure, adjacency matrices and advanced feature selection will be investigated as part of our future work.

%In this paper, we classify and analyze the adjacency matrix of the graph neural network and presented a novel transportation demand prediction model based on spatial-temporal attention mechanism and STGCN (AST-GCN). In particular, we propose a temporal attention module to extract temporal attention information, and an adaptive adjacency matrix is used to extract spatial attention information. Experiments were conducted on two different real-world dataset: Dublinbikes and NYC-citi dataset. Experiment shows that proposed AST-GCN can work better than most of exist methods in the NYC-citi dataset, and the prediction result on Dublinbikes dataset achieves 1.00 MAE. This work result has important guiding significance for how we choose the adjacency matrix for spatial-temporal demand prediction problems. Furthermore, this research represented a new perspective in graph convolutional network combine with attention mechanism. In the future, we will study the network structure and adjacency matrix that more closely integrate the attention mechanism and temporal and spatial relationships.

\vspace{-0.1cm}
\section*{Acknowledgment}
This work is supported in part by Science Foundataion Ireland SFI Grant SFI/12/RC/2289 P2. Hongde Wu is supported by the research master scholarship at Dublin City University.

\bibliographystyle{ieeetr}
\bibliography{References}

\begin{thebibliography}{10}

\bibitem{otero2018health}
I.~Otero, M.~Nieuwenhuijsen, and D.~Rojas-Rueda, ``Health impacts of bike
  sharing systems in europe,'' {\em Environment international}, vol.~115,
  pp.~387--394, 2018.

\bibitem{raviv2013optimal}
T.~Raviv and O.~Kolka, ``Optimal inventory management of a bike-sharing
  station,'' {\em Iie Transactions}, vol.~45, no.~10, pp.~1077--1093, 2013.

\bibitem{shi2015convolutional}
X.~Shi, Z.~Chen, H.~Wang, D.~Y. Yeung, W.~K. Wong, and W.~C. Woo,
  ``Convolutional lstm network: A machine learning approach for precipitation
  nowcasting,'' {\em Advances in neural information processing systems},
  vol.~2015, pp.~802--810, 2015.

\bibitem{wu2016short}
Y.~Wu and H.~Tan, ``Short-term traffic flow forecasting with spatial-temporal
  correlation in a hybrid deep learning framework,'' {\em arXiv preprint
  arXiv:1612.01022}, 2016.

\bibitem{DBLP:conf/ijcai/YuYZ18}
B.~Yu, H.~Yin, and Z.~Zhu, ``Spatio-temporal graph convolutional networks: {A}
  deep learning framework for traffic forecasting,'' in {\em Proceedings of the
  Twenty-Seventh International Joint Conference on Artificial Intelligence,
  {IJCAI} 2018, July 13-19, 2018, Stockholm, Sweden}, pp.~3634--3640,
  ijcai.org, 2018.

\bibitem{DBLP:conf/nips/DefferrardBV16}
M.~Defferrard, X.~Bresson, and P.~Vandergheynst, ``Convolutional neural
  networks on graphs with fast localized spectral filtering,'' in {\em Advances
  in Neural Information Processing Systems 29: Annual Conference on Neural
  Information Processing Systems 2016, December 5-10, 2016, Barcelona, Spain},
  pp.~3837--3845, 2016.

\bibitem{bai2019stg2seq}
L.~Bai, L.~Yao, S.~Kanhere, X.~Wang, Q.~Sheng, {\em et~al.}, ``Stg2seq:
  Spatial-temporal graph to sequence model for multi-step passenger demand
  forecasting,'' {\em arXiv preprint arXiv:1905.10069}, 2019.

\bibitem{zhang2017deep}
J.~Zhang, Y.~Zheng, and D.~Qi, ``Deep spatio-temporal residual networks for
  citywide crowd flows prediction,'' in {\em Proceedings of the AAAI Conference
  on Artificial Intelligence}, vol.~31, 2017.

\bibitem{yao2018deep}
H.~Yao, F.~Wu, J.~Ke, X.~Tang, Y.~Jia, S.~Lu, P.~Gong, J.~Ye, and Z.~Li, ``Deep
  multi-view spatial-temporal network for taxi demand prediction,'' in {\em
  Proceedings of the AAAI Conference on Artificial Intelligence}, vol.~32,
  2018.

\bibitem{chu2020passenger}
J.~Chu, X.~Wang, K.~Qian, L.~Yao, F.~Xiao, J.~Li, and Z.~Yang, ``Passenger
  demand prediction with cellular footprints,'' {\em IEEE Transactions on
  Mobile Computing}, 2020.

\bibitem{yao2019revisiting}
H.~Yao, X.~Tang, H.~Wei, G.~Zheng, and Z.~Li, ``Revisiting spatial-temporal
  similarity: A deep learning framework for traffic prediction,'' in {\em
  Proceedings of the AAAI conference on artificial intelligence}, vol.~33,
  pp.~5668--5675, 2019.

\bibitem{velivckovic2017graph}
P.~Veli{\v{c}}kovi{\'c}, G.~Cucurull, A.~Casanova, A.~Romero, P.~Lio, and
  Y.~Bengio, ``Graph attention networks,'' {\em arXiv preprint
  arXiv:1710.10903}, 2017.

\bibitem{chen2020multitask}
Z.~Chen, B.~Zhao, Y.~Wang, Z.~Duan, and X.~Zhao, ``Multitask learning and
  gcn-based taxi demand prediction for a traffic road network,'' {\em Sensors},
  vol.~20, no.~13, p.~3776, 2020.

\bibitem{kim2019graph}
T.~S. Kim, W.~K. Lee, and S.~Y. Sohn, ``Graph convolutional network approach
  applied to predict hourly bike-sharing demands considering spatial, temporal,
  and global effects,'' {\em PloS one}, vol.~14, no.~9, p.~e0220782, 2019.

\bibitem{ye2020coupled}
J.~Ye, L.~Sun, B.~Du, Y.~Fu, and H.~Xiong, ``Coupled layer-wise graph
  convolution for transportation demand prediction,'' {\em arXiv preprint
  arXiv:2012.08080}, 2020.

\bibitem{wu2019graph}
Z.~Wu, S.~Pan, G.~Long, J.~Jiang, and C.~Zhang, ``Graph wavenet for deep
  spatial-temporal graph modeling,'' in {\em International Joint Conference on
  Artificial Intelligence 2019}, pp.~1907--1913, Association for the
  Advancement of Artificial Intelligence (AAAI), 2019.

\bibitem{chiang2019cluster}
W.-L. Chiang, X.~Liu, S.~Si, Y.~Li, S.~Bengio, and C.-J. Hsieh, ``Cluster-gcn:
  An efficient algorithm for training deep and large graph convolutional
  networks,'' in {\em Proceedings of the 25th ACM SIGKDD International
  Conference on Knowledge Discovery \& Data Mining}, pp.~257--266, 2019.

\bibitem{shiraki2020spatial}
K.~Shiraki, T.~Hirakawa, T.~Yamashita, and H.~Fujiyoshi, ``Spatial temporal
  attention graph convolutional networks with mechanics-stream for
  skeleton-based action recognition,'' in {\em Proceedings of the Asian
  Conference on Computer Vision}, 2020.

\bibitem{zhang2020sta}
W.~Zhang, Z.~Lin, J.~Cheng, C.~Ma, X.~Deng, and H.~Wang, ``Sta-gcn: two-stream
  graph convolutional network with spatial--temporal attention for hand gesture
  recognition,'' {\em The Visual Computer}, vol.~36, no.~10, pp.~2433--2444,
  2020.

\bibitem{2016DNN}
J.~Zhang, Y.~Zheng, D.~Qi, R.~Li, and X.~Yi, ``Dnn-based prediction model for
  spatio-temporal data,'' in {\em the 24th ACM SIGSPATIAL International
  Conference}, 2016.

\bibitem{xgboost2016}
T.~Chen and C.~Guestrin, ``Xgboost: A scalable tree boosting system,'' in {\em
  Proceedings of the 22nd ACM SIGKDD International Conference on Knowledge
  Discovery and Data Mining}, KDD '16, (New York, NY, USA), p.~785–794,
  Association for Computing Machinery, 2016.

\bibitem{lstm1997}
S.~Hochreiter and J.~Schmidhuber, ``Long short-term memory,'' vol.~9,
  p.~1735–1780, Nov. 1997.

\bibitem{2017arXiv170701926L}
Y.~{Li}, R.~{Yu}, C.~{Shahabi}, and Y.~{Liu}, ``{Diffusion Convolutional
  Recurrent Neural Network: Data-Driven Traffic Forecasting},'' {\em arXiv
  e-prints}, p.~arXiv:1707.01926, July 2017.

\bibitem{kingma2017adam}
D.~P. Kingma and J.~Ba, ``Adam: A method for stochastic optimization,'' 2017.

\end{thebibliography}

\vspace{12pt}
\end{document}